\documentclass[10pt,twocolumn,letterpaper]{article}

\usepackage{wacv}
\usepackage{times}
\usepackage{epsfig}
\usepackage{graphicx}
\usepackage{amsmath}
\usepackage{amssymb}
\usepackage{booktabs}

\usepackage{multirow}
\usepackage{graphicx} 
\usepackage{caption,subcaption}
\usepackage{float}

\usepackage[dvipsnames]{xcolor}

\usepackage{subfiles} 
\usepackage[normalem]{ulem}

\usepackage[accsupp]{axessibility}

\newcommand{\vect}[1]{{\bf #1}}

\newcommand{\revadd}[1]{\textcolor{black}{#1}}

\def\wacvPaperID{500} 

\wacvalgorithmstrack   

\wacvfinalcopy 

\ifwacvfinal
\usepackage[breaklinks=true,bookmarks=false]{hyperref}
\else
\usepackage[pagebackref=true,breaklinks=true,colorlinks,bookmarks=false]{hyperref}
\fi

\begin{document}

\title{Frame Interpolation for Dynamic Scenes with Implicit Flow Encoding}

\author{Pedro Figueirêdo\\
Texas A\&M University\\
{\tt\small pedrofigueiredo@tamu.edu}
\and
Avinash Paliwal\\ 
Texas A\&M University\\
{\tt\small avinashpaliwal@tamu.edu}
\and
Nima Khademi Kalantari\\ 
Texas A\&M University\\
{\tt\small nimak@tamu.edu}
}

\maketitle
\thispagestyle{empty}

\begin{abstract}
   In this paper, we propose an algorithm to interpolate between a pair of images of a dynamic scene. While in the past years significant progress in frame interpolation has been made, current approaches are not able to handle images with brightness and illumination changes, which are common even when the images are captured shortly apart. We propose to address this problem by taking advantage of the existing optical flow methods that are highly robust to the variations in the illumination. Specifically, using the bidirectional flows estimated using an existing pre-trained flow network, we predict the flows from an intermediate frame to the two input images. To do this, we propose to encode the bidirectional flows into a coordinate-based network, powered by a hypernetwork, to obtain a continuous representation of the flow across time. Once we obtain the estimated flows, we use them within an existing blending network to obtain the final intermediate frame. Through extensive experiments, we demonstrate that our approach is able to produce significantly better results than state-of-the-art frame interpolation algorithms.
\end{abstract}

\section{Introduction}

\begin{figure}
    \includegraphics[width=\linewidth]{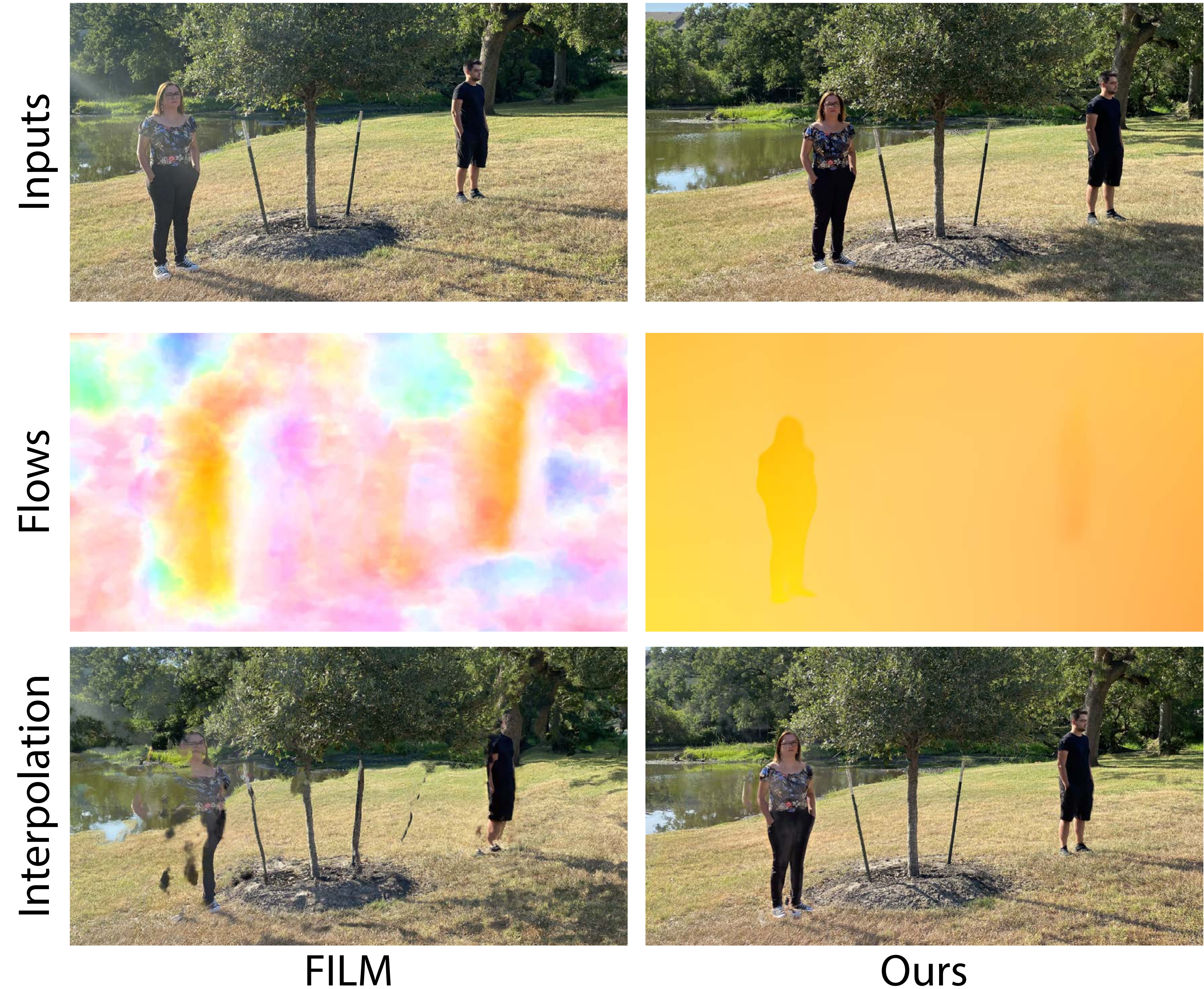}
    \vspace{-0.2in}
    \caption{We propose an approach to interpolate between two images of a dynamic scene. As shown on the top, the two images could often have significantly different lightings (even if taken shortly apart). Note the shadows on the ground in the two input images. Existing approaches, like Reda \etal~\cite{Reda_2019_CVPR} (FILM), train their system on datasets with no lighting variations. As a result, they are not able to effectively estimate the required flows and consequently generate final intermediate images with significant artifacts. We utilize \revadd{fixed} pre-trained flow networks that are highly robust to the illumination changes. As shown, our method is able to generate high-quality intermediate flows and results.}
    \label{fig:raft-film-flow}
    \vspace{-0.22in}
\end{figure}


With the widespread availability of smartphones featuring increasingly powerful cameras, taking professional high resolution photos has become a simple press of a button. Often, people take many pictures in search for the best representation of a moment in terms of the expression, pose, lighting, and exposure. Interpolating these photos creates a video with an exciting effect that provides an appealing way of remembering key moments.

Existing video interpolation approaches~\cite{Huang_2022_rife,Park_2021_ICCV} often struggle to handle these cases because of significant camera and scene motion. Recently, Reda \etal~\cite{reda2022film} propose to overcome this challenge by leveraging a multi-scale feature extraction and flow estimation strategy. Specifically, they use a series of convolutional layers to construct a feature pyramid and use it to estimate a set of two flows from the in-between frame to the two input images at multiple scales. These flows are then used to warp the two images along with their features at each scale to the frame of interest. The warped features and images are then aggregated and combined to produce the final frame.

While this approach produces interpolated results with high quality, it is limited to cases where the two images have consistent brightness and illumination. However, in practice, the two images could have different brightness and illumination even if taken shortly apart, as shown in Fig.~\ref{fig:raft-film-flow}. Unfortunately, the approach by Reda \etal~\cite{reda2022film} is not able to handle these cases, producing results with unnatural motion and severe artifacts. 

Our key observation is that the source of the problem is that their estimated flows degrade quickly in presence of illumination changes (see Fig.~\ref{fig:raft-film-flow}). This is largely because their network is trained on video datasets~\cite{xue2019video,sim2021xvfi} where the frames are captured in quick succession, and thus have consistent brightness. Therefore, the images with lighting variations fall outside the distribution of their training data. On the other hand, their blending, even though trained on video datasets, is usually able to generate pleasing intermediate images. The key \revadd{to generating high-quality images thus lies in improving the quality of the estimated flows.}

In this paper, we propose to address this problem by utilizing a pre-trained optical flow network. Existing optical flow methods~\cite{teed2020raft,zhang2021separable} are highly robust to even significant lighting changes, and thus are suitable for our application. The major challenge here is that these methods only estimate the flow between two images, but we need the flows from an intermediate frame to the two input images. 

To overcome this challenge, we propose a per-scene optimization method (no training on large datasets) by utilizing implicit neural networks~\cite{mildenhall2020nerf,sitzmann2020implicit}. Our key idea is that by encoding the bidirectional flows between the two input images into a coordinate-based network, we essentially obtain a continuous representation of flows across time. Therefore, we can use such a network to estimate the flows at any in-between time coordinate. To be able to properly estimate the intermediate flows, we use a hypernetwork that takes the time coordinate and estimates the weights of a coordinate-based neural network. The coordinate-based network then estimates the flow at each pixel by taking the pixel coordinate as the input. We optimize the hypernetwork using the bidirectional flows and then estimate any in-between flows by passing the appropriate time coordinate to this optimized hypernetwork. We then use these intermediate flows with Reda \etal's blending network to generate the final intermediate images.

We show that our method outperforms existing approaches on a wide range of challenging scenes with large lighting variations and motions (see Figs.~\ref{fig:raft-film-flow}~and~\ref{fig:results} and the supplementary video). Furthermore, we justify our design choices through extensive experiments.

\section{Related Work}

In this section, we review the frame interpolation methods, as well as the approaches in image morphing, a relevant but different problem. We also briefly discuss implicit neural representations as we utilize them in our work.

\subsection{Frame Interpolation} 

In recent years, deep learning methods have become popular due to their effectiveness in handling challenging scenarios like scenes with large complex motions. Niklaus and Liu~\cite{niklaus2018context} use a pretrained flow network to warp the existing frames and then use a context aware blending network to synthesize the interpolated frame. Similarly, Jiang \etal~\cite{jiang2018super} generate interpolated frames at arbitrary time by estimating the flows to the intermediate frame. Wenbo \etal~\cite{DAIN} utilize a depth estimation network to handle the occlusions. Niklaus and Liu~\cite{Niklaus_2020_CVPR} use forward warping with a synthesis network to generate interpolated frames. 

Moreover, Park et al.~\cite{Park2020bmbc} propose a model based on bilateral motion estimation to generate high quality warped frames for blending. They further enhance this approach~\cite{Park_2021_ICCV} by computing asymmetric bilateral fields to account for non-linearities in the scene. Huang et al.~\cite{Huang_2022_rife} directly estimate the intermediate flows by using a privileged distillation scheme during training. Reda \etal~\cite{reda2022film} propose a unified network consisting feature pyramid and flow extraction with fusion components to handle scenes with large motions. This method along with a few other recent approaches~\cite{lee2022enhanced,hu2022many} focus on improving the quality on high resolution videos. Furthermore, a couple of approaches~\cite{Liu2019deep,Reda_2019_CVPR} propose to improve the performance of the supervised methods by further training the system in an unsupervised manner.

In contrast to these methods, several approaches propose to directly estimate the final image without explicitly estimating the required flows. For example, Niklaus \etal~\cite{niklaus2017video,Niklaus2017videoframe} use adaptive convolution kernels to generate the intermediate frame from the neighboring images. Choi \etal~\cite{choi2020cain} use PixelShuffle~\cite{Shi2016real} with channel attention~\cite{Woo2018cbam} to directly synthesize the middle frame. Gui \etal~\cite{Gui2020featureflow} blend deep features and Kalluri \etal~\cite{kalluri2020flavr} utilize 3D space-time convolutions for interpolation. Unfortunately, all of these techniques, as well as the flow-based methods, train their system on sequences with consistent illumination, and thus are not suitable for our application.

Related to our work, Bemana \etal~\cite{bemana2020x} learn an implicit mapping between view-time-light coordinates and input images by optimizing a convolutional network. However, they use a loss between the input and warped images for optimization, and thus their main assumption is brightness constancy which is invalid in our cases. We address this problem by utilizing a pre-trained flow network that is highly robust to illumination changes.

\subsection{Morphing}
\label{sec:morphing}
Similar to our application, image morphing methods produce a series of images to smoothly transition between two input images. Most algorithms~\cite{Beier1992feature,Chen1993view,Seungyong1996image,Seitz1996view,Liao_2014_ToG} achieve this by first computing a set of sparse correspondences between the two images, and then using them to warp the images to an intermediate frame. These warped images are then combined to create a morphed image. These methods are typically best suited for images of different scenes. Since they typically utilize sparse correspondences, their transitions for our examples (images of the same dynamic scene) are not sufficiently detailed to produce appealing effects.

Several approaches~\cite{Shechtman_2010_CVPR,Darabi_2012_ToG} propose to handle this application using patch-based optimization systems. Similarly, these methods are suitable for different objects/scenes and are not able to produce visually pleasing result for images of the same object/scene. Recent breakthroughs in deep learning and generative adversarial networks~\cite{brock2018large,Karras2020analyzing} have enabled efficient and high quality morphing by interpolating in the latent space~\cite{Abdal2019image2stylegan,Jahanian2020On,Park2020neural}. However, these approaches mostly work for a single or few objects (e.g., faces, cats, cars) and are not general.

\subsection{Implicit Neural Representations}
\label{sec:implicitneuralrepr}
A large number of recent methods have used neural networks as a memory-efficient continuous function approximator for implicit representation of images~\cite{sitzmann2020implicit,tancik2020fourfeat,mueller2022instant,dupont2021coin,chen2020learning} and videos~\cite{hao2021nerv,zhang2022implicit,sitzmann2020implicit}, 3D objects via signed distance functions~\cite{Jiang_2020_CVPR,peng2020convolutional,Park2019deepsdf,Atzmon2020sal,sitzmann2019metasdf,Michalkiewicz2019implicit,sitzmann2020implicit} or occupancy networks~\cite{mescheder2019occupancy, chen2019learning}, and radiance fields~\cite{mildenhall2020nerf,kangle2021dsnerf,Jain_2021_ICCV,hedman2021snerg,nerv2021}. Novel input encodings~\cite{mildenhall2020nerf,tancik2020fourfeat} and activation functions~\cite{sitzmann2020implicit} have been instrumental in enabling the encoding of high frequency details in compact networks for these applications.  We build upon these advances, but use implicit neural representations for optical flow interpolation.

\begin{figure*}
    \includegraphics[width=\linewidth]{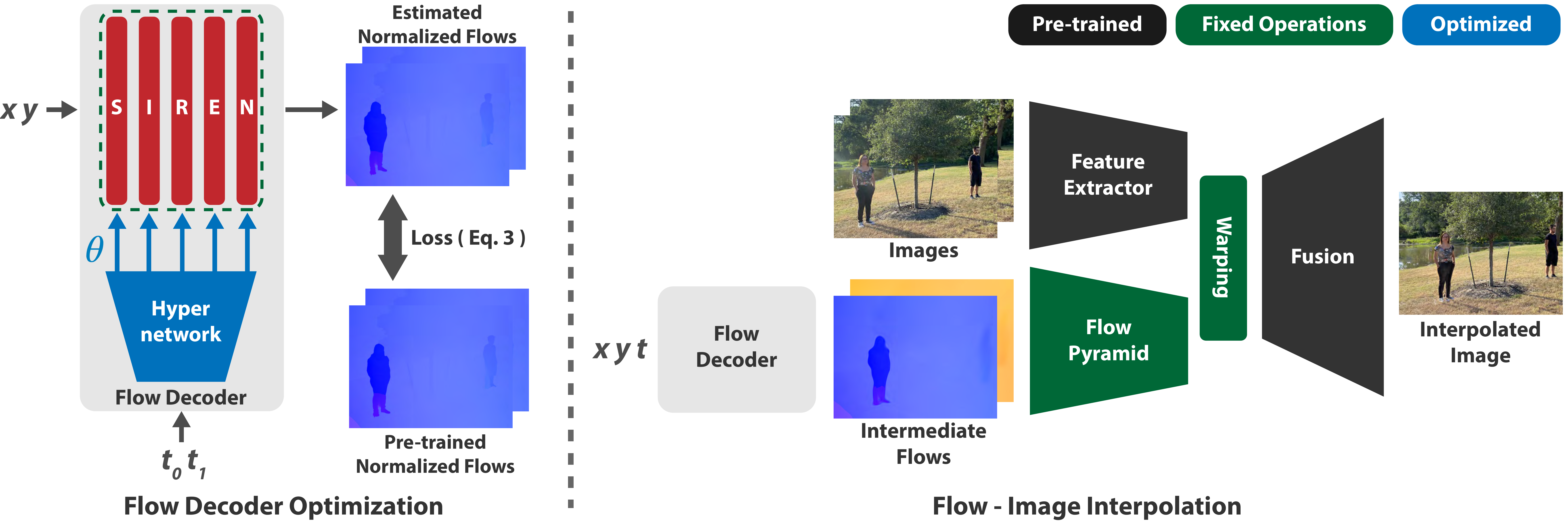}
    \vspace{-0.2in}
    \caption{We provide an overview of our approach. On the left, we \revadd{perform an optimization to} encode the bidirectional flows estimated using a pre-trained flow network into our system, consisting of a hypernetwork and a coordinate-based network (SIREN). Once the flows are encoded, we estimate the intermediate flows through our system and use these flows within FILM's blending system~\cite{Reda_2019_CVPR} to generate the final image, as shown on the right. \revadd{Here, ``Pre-trained'' means the components are trained on a large dataset in an offline manner and are fixed at run-time. ``Fixed-Operations'' refers to untrainable and fixed downsampling (flow pyramid) and bilinear interpolation (warping). Finally, ``Optimized'' indicates that the component (hypernetwork) is trained at run-time on the example at hand.}}
    \label{fig:overview}
    \vspace{-0.2in}
\end{figure*}

\section{Method}
\label{sec:method}

Given a pair of images of a dynamic scene, $I_{t_0}$ and $I_{t_1}$, captured under different conditions, e.g., different exposures, the goal of our method is to reconstruct an image at time $t$ between the two images, where  $t \in [t_0,t_1]$. Most existing frame interpolation methods, and in particular the state-of-the-art method of Reda \etal~\cite{reda2022film}, break down this process into flow estimation and blending components. Specifically, they first compute a set of flows from the intermediate frame at time $t$ to the two input images, $F_{t \rightarrow t_0}$ and $F_{t \rightarrow t_1}$. They then use these flows to backward warp the images/features to the intermediate frame and combine them to reconstruct the final image $I_t$.

Unfortunately, these approaches are not able to properly handle cases with illumination variation mainly because the quality of the estimated flows degrades quickly in absence of brightness constancy. This is expected as these methods train their system on video datasets that contain minimal lighting variations between the neighboring frames.

To address this problem, we propose to utilize the powerful optical flow estimation methods~\cite{teed2020raft,zhang2021separable} that are highly robust to these illumination changes. The major challenge is that using these optical flow methods, we can only estimate the bidirectional flows between the inputs, $F_{t_0 \rightarrow t_1}$ and $F_{t_1 \rightarrow t_0}$, but we require estimating the intermediate flows. We propose to address this challenge by implicitly interpolating the bidirectional optical flows using a coordinate-based neural network. Once the intermediate flows are estimated, we incorporate them in the blending network by Reda~\etal~\cite{reda2022film} to estimate the final image. The overview of our system is shown in Fig.~\ref{fig:overview}. Next, we discuss our implicit flow interpolation and blending approaches.

\vspace{-0.15in}
\paragraph{Discussion:} One might attempt to train existing video interpolation methods on datasets with lighting variations to handle this application. However, constructing a dataset of input images with realistic illumination changes and their corresponding intermediate ground truth image is difficult. Moreover, even if the dataset can be reconstructed, it might be challenging to design a network that can outperform state-of-the-art optical flow methods. Finally, by using existing flow estimation methods, we have the ability to replace them with newer and better methods to further improve our results in the future. We also note that we considered forward warping, instead of backward warping, using Niklaus \etal's method~\cite{Niklaus_2020_CVPR} to avoid the need for computing the intermediate flows. However, their approach requires computing an importance mask to properly handle the overlapping regions. Unfortunately, this importance mask is computed with the assumption of brightness constancy which is invalid in our cases.

\subsection{Implicit Flow Interpolation}

Our goal here is to estimate the intermediate flows, $F_{t \rightarrow t_0}$ and $F_{t \rightarrow t_1}$, from the bidirectional flows between the two input images, $F_{t_0 \rightarrow t_1}$ and $F_{t_1 \rightarrow t_0}$, generated by an existing pre-trained flow network. In our system, we use the optical flow method of Teed~\etal~\cite{teed2020raft} because of its ability to generate high quality flows in challenging cases. As discussed, we propose to estimate the intermediate flows implicitly through a coordinate-based neural network.

A coordinate-based network finds a mapping between an input coordinate and the corresponding output at that coordinate, i.e., $\vect{y} = f_\theta(\vect{x})$, where $\theta$ refers to the weights of the network. This network can then be optimized on a set of input output pairs $(\vect{x}_i, \vect{y}_i)$ by minimizing a loss to find optimal network weights $\theta$. Through this optimization, the data will be encoded into the weights of the neural network. The key idea is that, once this optimization is performed, we can evaluate the network at any in-between coordinate to obtain the corresponding output. The network will essentially interpolate the output at the observed coordinates to generate the intermediate results.

In our application, the input coordinates are 3-dimensional, $\vect{x} = (x, y, t)$, where $x$ and $y$ are the spatial and $t$ is the time coordinate. The output, on the other hand, is a 2D flow (in horizontal and vertical directions) at each coordinate $\vect{y} = F(x, y, t)$. Note that we use flows that are normalized by the difference in the time coordinates as the output to our network, i.e., $F(x, y, t_0) = F_{t_0 \rightarrow t_1}(x, y) / (t_1 - t_0)$ and $F(x, y, t_1) = F_{t_1 \rightarrow t_0}(x, y) / (t_0 - t_1)$. This is because the original flows are in opposite directions ($t_0$ to $t_1$ and $t_1$ to $t_0$) and cannot be interpolated. By normalizing the flows with their coordinate difference the direction of the two flows become consistent. We encode these normalized flows into our coordinate-based network using the following objective:

\vspace{-0.25in}
\begin{equation}
    \label{eq:siren_loss}
    \theta^* = \arg \min_\theta \sum_{i = 1}^w  \sum_{j = 1}^h \sum_{k = 0}^1 \Vert f_\theta(x_i,y_j,t_k) - F(x_i,y_j,t_k) \Vert_2,
\end{equation}
\vspace{-0.2in}

\noindent where $w$ and $h$ are the width and height of the images (and similarly the flows), respectively.

Once this optimization is performed, we can evaluate the network at any arbitrary in-between time coordinate $t \in [t_0, t_1]$ to estimate the intermediate flow. Note that the network produces a normalized flow at each coordinate. We then convert this to the two intermediate flows as follows:

\vspace{-0.2in}
\begin{align}
    \label{eq:inter_flows}
     F_{t \rightarrow t_0}(x, y) &= (t - t_0) \times f_\theta(x, y, t) \nonumber\\ \revadd{F_{t \rightarrow t_1}(x, y)} &= (t-t_1) \times f_\theta(x, y, t).
\end{align}
\vspace{-0.2in}

\begin{figure}
    \includegraphics[width=\linewidth]{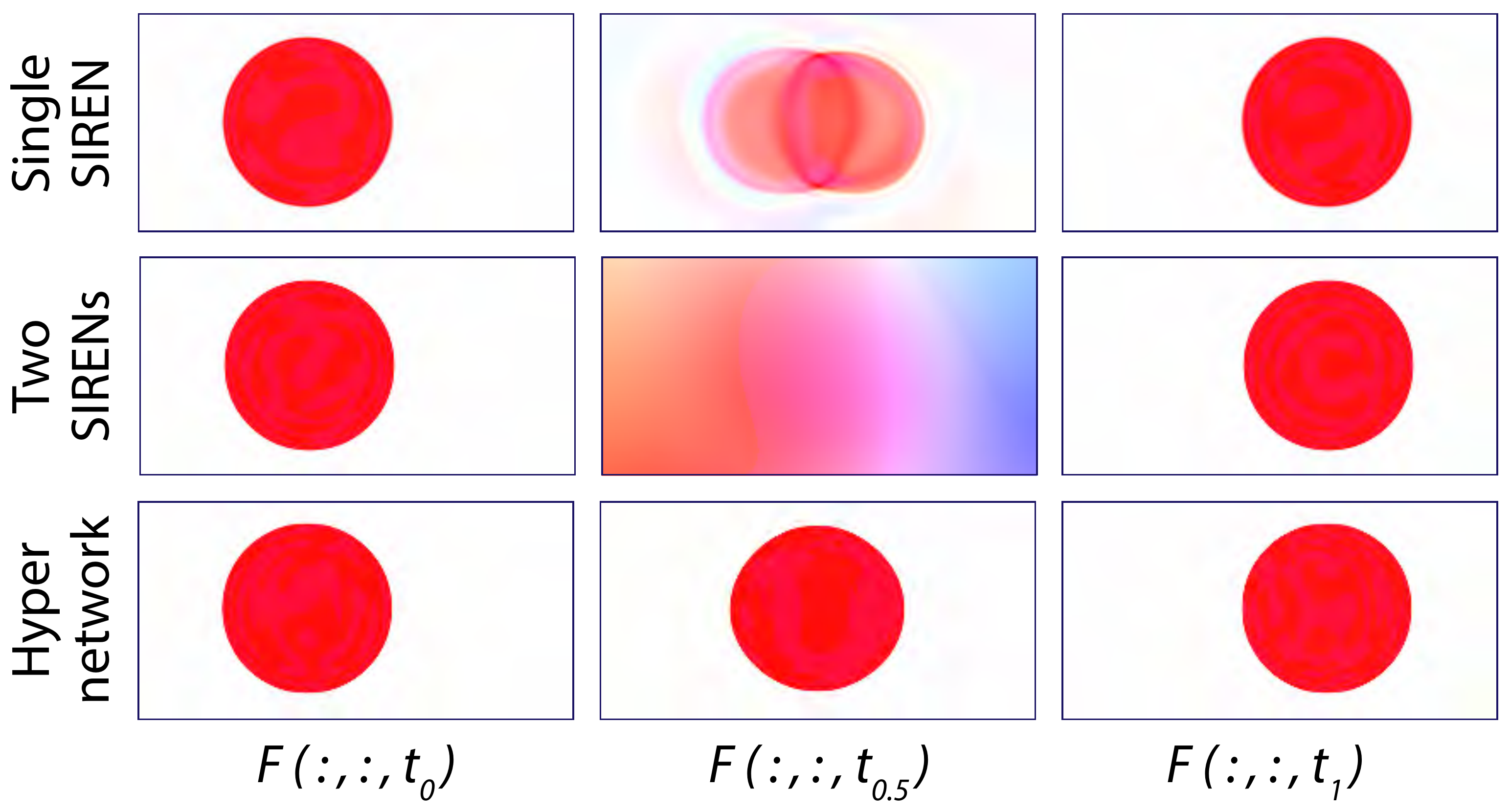}
    \vspace{-0.25in}
    \caption{We demonstrate the effectiveness of our approach through a synthetic example. Here, a circle is moving from left to right, and thus the flows at the two input coordinates, $t_0$ and $t_1$, contain a circle at different positions. From each case, we show the encoded input flows and the interpolated flow at time $t_{0.5}$. By encoding both flows into a single SIREN, the network simply produces an ``average'' of the two flows. Encoding the flows into two \revadd{ independent} SIRENs\revadd{, without an underlying hypernetwork,} and then interpolating their weights is not effective as the two flows are encoded separately. Using a hypernetwork, we are able to correctly produce the interpolated flow.}
    \label{fig:hypernohyper}
    \vspace{-0.2in}
\end{figure}

For our network, we use SIREN, as proposed by Sitzmann~\etal~\cite{sitzmann2019scene}, with 5 hidden layers each containing 128 neurons. Moreover, we set the frequency of the sinosuidal activation functions to 10. As shown in Fig.~\ref{fig:hypernohyper}, this approach (Single SIREN) is not able to properly interpolate the two input flows. This is because a coordinate-based network, generates the in-between results by ``averaging'' the data at the observed nearby coordinates. Therefore, the network reconstructs the intermediate flow by combining the two flows ($t_0$ and $t_1$) at the same spatial coordinates. Essentially, the network generates the ``average'' of the two encoded flows resulting in blurry intermediate flows. We address this problem by encoding the time coordinate using a hypernetwork, as discussed next.

\subsection{Time Encoding Through A Hypernetwork}

To properly interpolate the intermediate flow, we need to ``average'' the shape of the two input flows. Our main observation is that by encoding data into a coordinate-based network, the shape will be essentially represented using the weights of the network. Therefore, we can encode the normalized bidirectional flows into two separate coordinate-based networks. \revadd{The networks in this case, only takes the spatial coordinates as the input $f_\theta(x_i, y_i)$ and the two networks are independently optimized to encode the flows at $t_0$ and $t_1$.} The optimized weights, $\theta_0$ and $\theta_1$, can then be linearly interpolated to obtain the representation of the intermediate flow $\theta_t$. The interpolated weights can be used to generate the in-between flow. However, as shown in Fig.~\ref{fig:hypernohyper}, this strategy (Two SIRENs) does not produce the desired interpolation as the two flows are independently encoded and the two representations, $\theta_0$ and $\theta_1$, are not interpolatable.

To address this problem, we propose to estimate the weights of the coordinate-based network using a hypernetwork~\cite{ha2016hypernetworks} that takes the time coordinate as the input, i.e., $\theta = f_\phi(t)$. We encode both normalized flows into our system through the following objective:

\vspace{-0.2in}
\begin{align}
    \label{eq:hyper_loss}
    \phi^* = \arg \min_\phi \sum_{i = 1}^w  &\sum_{j = 1}^h \sum_{k = 0}^1 \Vert f_\theta(x_i, y_j) - F(x_i, y_j, t_k) \Vert_2, \nonumber \\
    &\text{where} \quad \theta = f_\phi(t_k).
\end{align}
\vspace{-0.2in}

In this case, we encode the two flows simultaneously and since the representations (weights $\theta_0 = f_\phi(t_0)$ and $\theta_1 = f_\phi(t_1)$) are estimated using a single hypernetwork, they are closer together in the high dimensional space, and thus are interpolatable. \revadd{The main difference with respect to Two SIRENs strategy is that, here, the estimated representations (weights) are produced by the same network (hypernetwork). While, in theory, $\phi^*$ that produces the same $\theta_0$ and $\theta_1$ as in the two SIREN strategy is a valid minimizer of Eq.~\ref{eq:hyper_loss}, by using a small hypernetwork and initializing the weights to small values, we usually converge to a solution that produces highly correlated SIREN weights in practice.} 

\revadd{Once the hypernetwork is optimized, we generate the final intermediate flows by evaluating the hypernetwork at the in-between time coordinate $\theta_t = f_\phi(t)$ and using the calculated weights in Eq.~\ref{eq:inter_flows}. We also experimented with first estimating the weights at $t_0$ and $t_1$ using our hypernetwork (i.e., $\theta_0 = f_\phi(t_0)$ and $\theta_1 = f_\phi(t_1)$) and then linearly interpolating them to obtain $\theta_t$, but the results were similar.}

Our hypernetwork is composed of a set of fully-connected networks with one hidden layer of size 128 with ReLU activation that maps the time coordinate $t$ to the weights in each layer of our coordinate-based network (SIREN). \revadd{Moreover, we use $t_0 = 0$ and $t_1 = 0.1$ to further force the network to produce highly correlated weights (see the effect of coordinate distance in Fig.~\ref{fig:coordablation}).}

\revadd{In summary, using the small hypernetwork along with the close time coordinates, we confine the space of possible weights, which is essential for high-quality interpolation.} As shown in Fig.~\ref{fig:hypernohyper}, our system with a hypernetwork is able to properly encode the two flows and reconstruct an intermediate flow.

\subsection{Blending}

As discussed, we use our interpolated flows with the pre-trained blending network by Reda~\etal~\cite{reda2022film} (FILM) to generate the final interpolated images. While FILM's blending system is trained on standard datasets with mostly constant illumination, we observe that it produces visually pleasing interpolation between images with varying lighting. To incorporate our estimated intermediate flows into their system, we first generate a flow pyramid by downsampling our estimated intermediate flows to multiple scales. We use bilinear interpolation to downsample the flows and divide the magnitude of the flows by the scale factor. Once we obtain the pyramid of the two flows, we use them to warp the feature pyramid (calculated with FILM's feature extractor), as well as the input images, to the intermediate frame. Finally, we pass all the warped features and images to FILM's fusion network to obtain the final result.

\section{Results}

We implement our model in PyTorch \cite{NEURIPS2019_9015} and utilize Torchmeta~\cite{deleu2019torchmeta} for our hypernetwork. We leverage pre-trained flow estimation network of Teed~\etal~\cite{teed2020raft} (RAFT) and the blending network by Reda~\etal~\cite{reda2022film} (FILM). Our solution uses the \emph{sintel} checkpoint for RAFT and the \emph{style} checkpoint for FILM's blending. \revadd{Specifically, RAFT's \emph{sintel} checkpoint has been trained on pairs of images with their corresponding ground truth flow from the Sintel dataset~\cite{janai2017slow}. RAFT applies various data augmentations to ensure robustness to a variety of distortions.} We optimize our model using Adam~\cite{kingma2014adam}, with the default parameters $\beta_1=0.9$ and $\beta_2=0.999$. We train for 10K iterations on a single A100 GPU using a learning rate of $1e^{-6}$.

\subsection{\revadd{Comparisons}}

\revadd{We compare our algorithm to state-of-the-art video frame interpolation approaches by Park~\etal~\cite{Park_2021_ICCV} (ABME) and Reda~\etal~\cite{reda2022film} (FILM). We use the source code provided by the authors for both approaches.}

\vspace{-0.3in}
\revadd{\paragraph{Quantitative:} Numerical evaluation of the quality of the interpolated images is challenging as there are no datasets containing input images with lighting variations and their corresponding ground truth intermediate images. While we could potentially use existing datasets and apply various perturbations (e.g., hue) to the input images, constructing the corresponding ground truth intermediate image remains a challenge as these perturbations are non-linear.}

\revadd{
    Therefore, we only numerically evaluate the quality of the intermediate flows. To do so, we use the two video frame interpolation datasets of Xiph 2K and 4K~\cite{Niklaus_2020_CVPR}, as well as the Sintel dataset~\cite{janai2017slow}. For each input image pair, we apply various photometric augmentations by randomly perturbing brightness, contrast, saturation, and hue. We use Pytorch's \textsc{ColorJitter} with brightness 0.4, contrast 0.4, saturation 0.4, and hue 0.5/$\pi$. We then use the perturbed images as the input to estimate the intermediate flows which are then compared to the reference flows. The reference intermediate flows for the Sintel dataset are provided, but for Xiph 2K and 4K we simply use the RAFT flows between the intermediate and two input clean (unperturbed) images as the reference. For Xiph 2k and 4k, we skip six frames when creating the image pairs (e.g. 1-7, 2-8, etc.) to increase the amount of motion of the scenes, while we skip one frame for the Sintel dataset.}

\revadd{
    Table \ref{tab:ResultsView} shows the comparison against the approach by Reda~\etal~\cite{reda2022film} in terms of average end-point error (EPE). Note that we do not include the method by Park~\etal~\cite{Park_2021_ICCV} since their approach does not explicitly estimate flows. As seen, our interpolated flows are significantly better than the estimated intermediate flows by Reda~\etal. We show visual comparisons on a few images from all datasets in the supplementary material.
}

\begin{table}[!t]
\renewcommand{\arraystretch}{1.3}
\caption{We numerically compare our interpolated flows against the ones by Reda \etal~\cite{reda2022film} in terms of average end-point-error (EPE). For these comparisons we use Xiph 2K and 4K~\cite{Niklaus_2020_CVPR}, as well as Sintel~\cite{janai2017slow}.}
\vspace{-0.1in}
\centering
\begin{footnotesize}
\begin{tabular}{l|c|c|c}
  \hline\hline
  & Xiph 2K & Xiph 4K & Sintel\\
  \hline
  FILM & 13.97 & 37.34 & 12.9 \\
  Ours & \textbf{3.4} & \textbf{17.46} & \textbf{5.16}\\
  \hline\hline
\end{tabular}
\label{tab:ResultsView}
\end{footnotesize}
\vspace{-0.2in}
\end{table}

\vspace{-0.15in}
\paragraph{\revadd{Qualitative:}}

\begin{figure*}
\includegraphics[width=\linewidth]{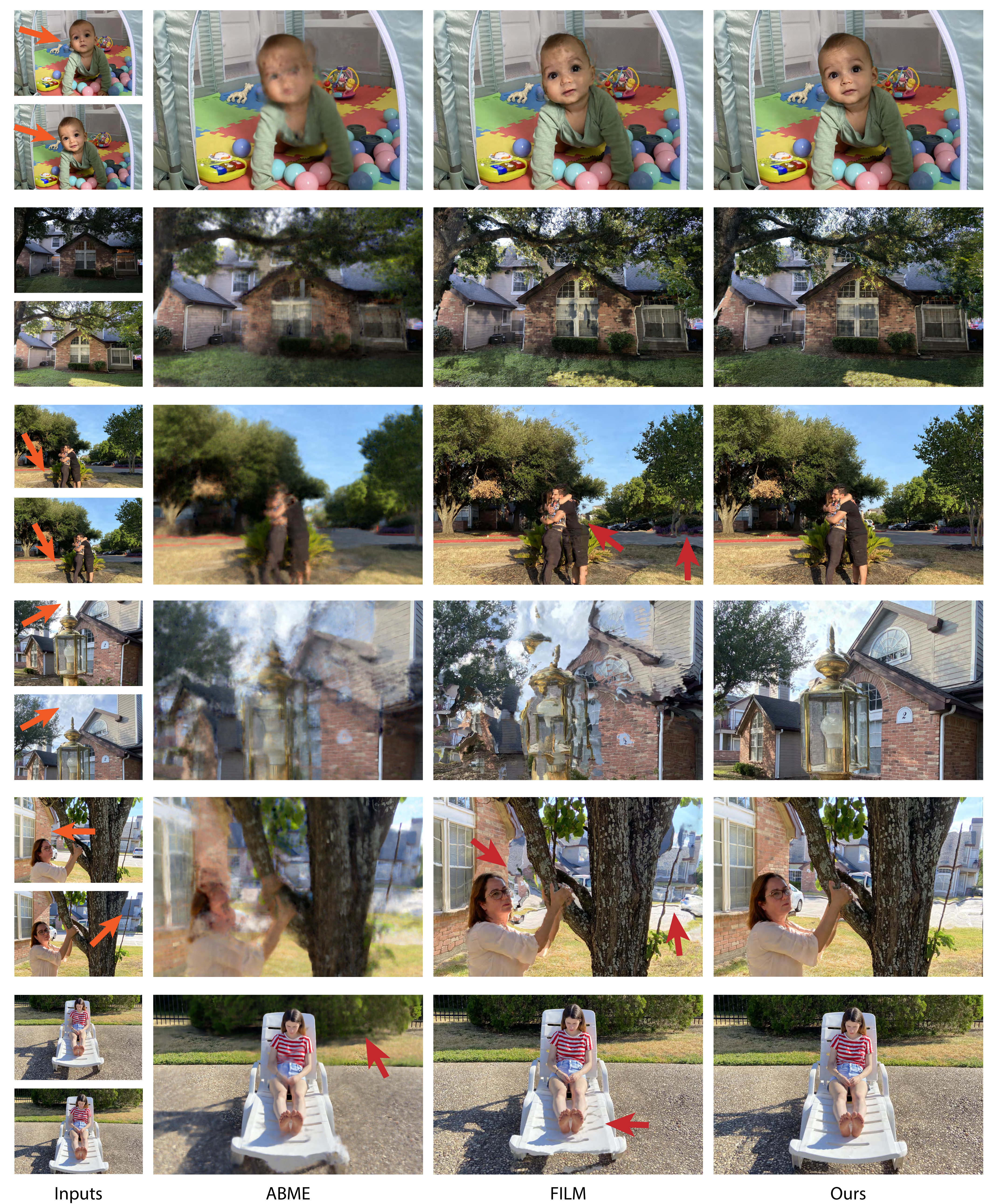}
\caption{We show comparisons against the state-of-the-art methods by Park \etal~\cite{Park_2021_ICCV} (ABME) and Reda \etal~\cite{Reda_2019_CVPR} (FILM).}
\label{fig:results}
\end{figure*}

We perform \revadd{qualitative} comparisons on several challenging scenes captured using a smartphone at resolution $2016\times 1512$. To show the robustness of our approach, we capture both indoor and outdoor scenes with image pairs taken at different times of the day or seconds apart. We compare the estimated middle image by all the approaches on 6 scenes in Fig.~\ref{fig:results}, but encourage the readers to see our supplementary video.

We begin by examining the \textsc{Baby} scene which contains non-rigid motion in an indoor setup. The baby transitions from a shaded curious pose to a partially lit smiley expression. This scene illustrates how natural lighting can change significantly even for photos taken seconds apart. ABME produces blurry results in the moving regions. While FILM generates a sharper interpolation, it deforms the baby's head. Our method preserves the baby's appearance, producing a realistic intermediate image in terms of expression, motion, and lighting. 

The \textsc{House} scene demonstrates our method's ability of interpolating images under extremely different lighting conditions. The two images are captured from a mostly static scene, but at different times of the day (morning and evening). ABME generates a blurry interpolation with ghosting artifacts, while FILM produces an unnatural interpolation by introducing dark patches throughout the image. Our method, on the other hand, is able to interpolate intermediate images with reasonable quality because of the ability of our system to interpolate high quality flows.

The \textsc{Hug} scene contains both moving subjects and significant camera motions. The slight lighting variations on the trees and shadows combined with the large motions make this scene extremely challenging for the other approaches. In contrast, RAFT is able to estimate high quality flows and our approach properly interpolates the intermediate flows to produce results without objectionable artifacts. Similarly, the \textsc{Lamp} scene, while static, contains significant camera motion and has been captured with different exposures (see the sky in the input images) making it a challenging scene for the other methods. Although our approach produces slight ghosting artifacts around the object boundaries, our results are still plausible and significantly better than the other techniques.

The \textsc{Tree} scene contains significant subject motions and lighting variations (see the building roof and shadows in the input). ABME blurs out the entire frame, while FILM severely distorts the background. In contrast, our approach produces a high-quality interpolation by smoothly warping the subject while maintaining a coherent background. Finally, although the \textsc{Lady} is a relatively easy scene, ABME produces a blurry background and FILM is not able to properly reconstruct the gaps in the chair. Our approach, however, produces a high quality results without any objectionable artifacts.

\subsection{Ablation Experiments}

\begin{figure}
\includegraphics[width=\linewidth]{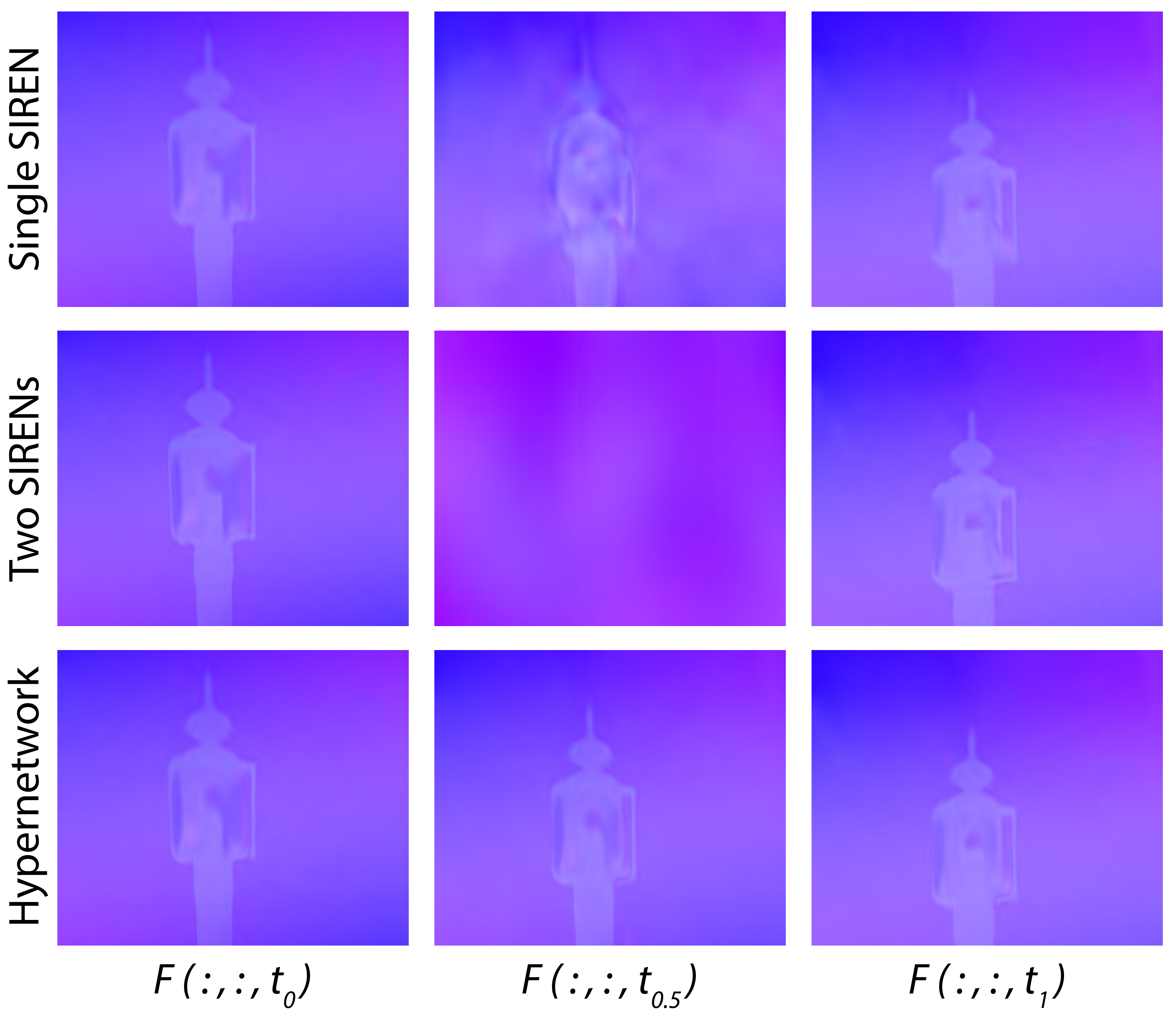}
\vspace{-0.27in}
\caption{We evaluate the impact of the hypernetwork on the quality of the interpolated flow on a real example. All the approaches properly encode the flows at $t_0$ and $t_1$, but only our approach with the hypernetwork can produce an interpolated flow with high quality.}
\label{fig:hypernetwork}
\vspace{-0.15in}
\end{figure}

\paragraph{Effect of Hypernetwork:} We begin by evaluating the effect of the hypernetwork on the quality of the interpolated flows. In Fig.~\ref{fig:hypernohyper}, we show its effect on a synthetic example. The effect on a real example is shown in Fig.~\ref{fig:hypernetwork}. As seen, while all the approaches are able to encode the input flows at times $t_0$ and $t_1$ with similar quality, only our method with a hypernetwork is able to properly reconstruct the interpolated flow.

\begin{figure}
    \includegraphics[width=\linewidth]{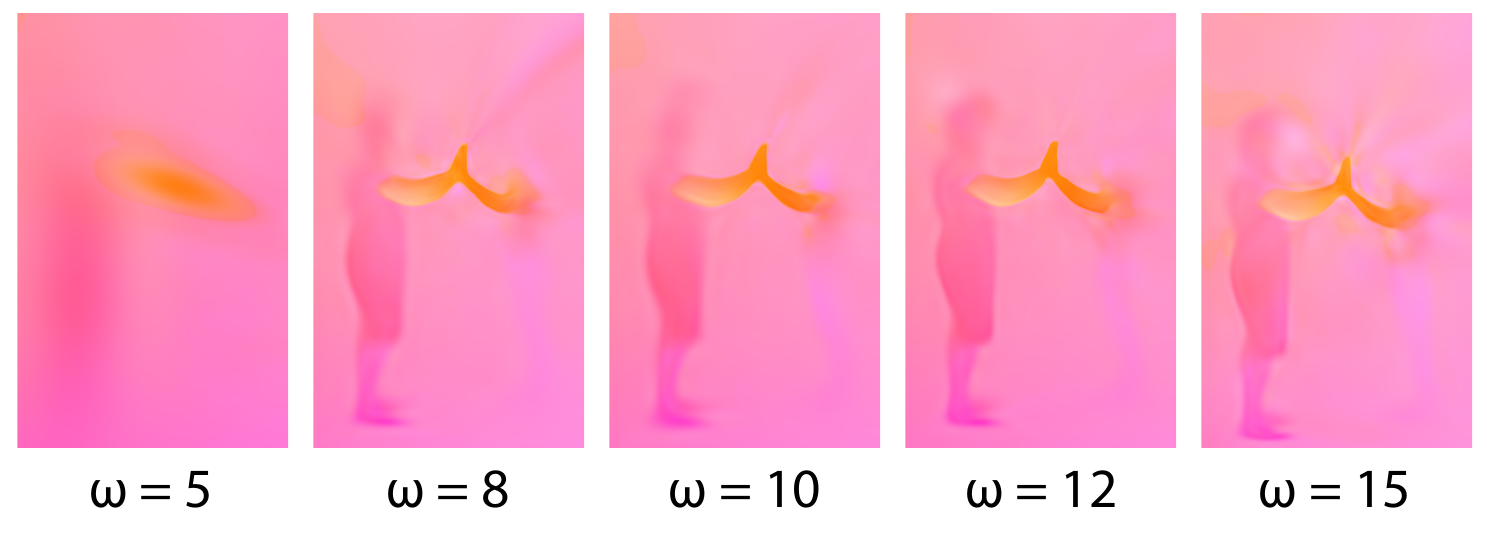}
    \vspace{-0.27in}
        \caption{Effect of varying $\omega$ of SIREN on the interpolated flow at $t_{0.5}$.}
        \label{fig:omegablation}
        \vspace{-0.2in}
\end{figure}

\vspace{-0.15in}
\paragraph{Effect of $\omega$:} Next, we evaluate the effect of the frequency $\omega$ of the sinusoidal activation functions of SIREN in Fig.~\ref{fig:omegablation}. The frequency is a key factor in properly encoding the data into a coordinate-based network. Higher frequencies are more suitable for signals containing a lot of details, while lower frequencies are more appropriate for smooth signals. As shown, frequencies between 8 to 12 produce reasonable results in our cases. We use $\omega = 10$ in our implementation. 

\begin{figure}
    \includegraphics[width=\linewidth]{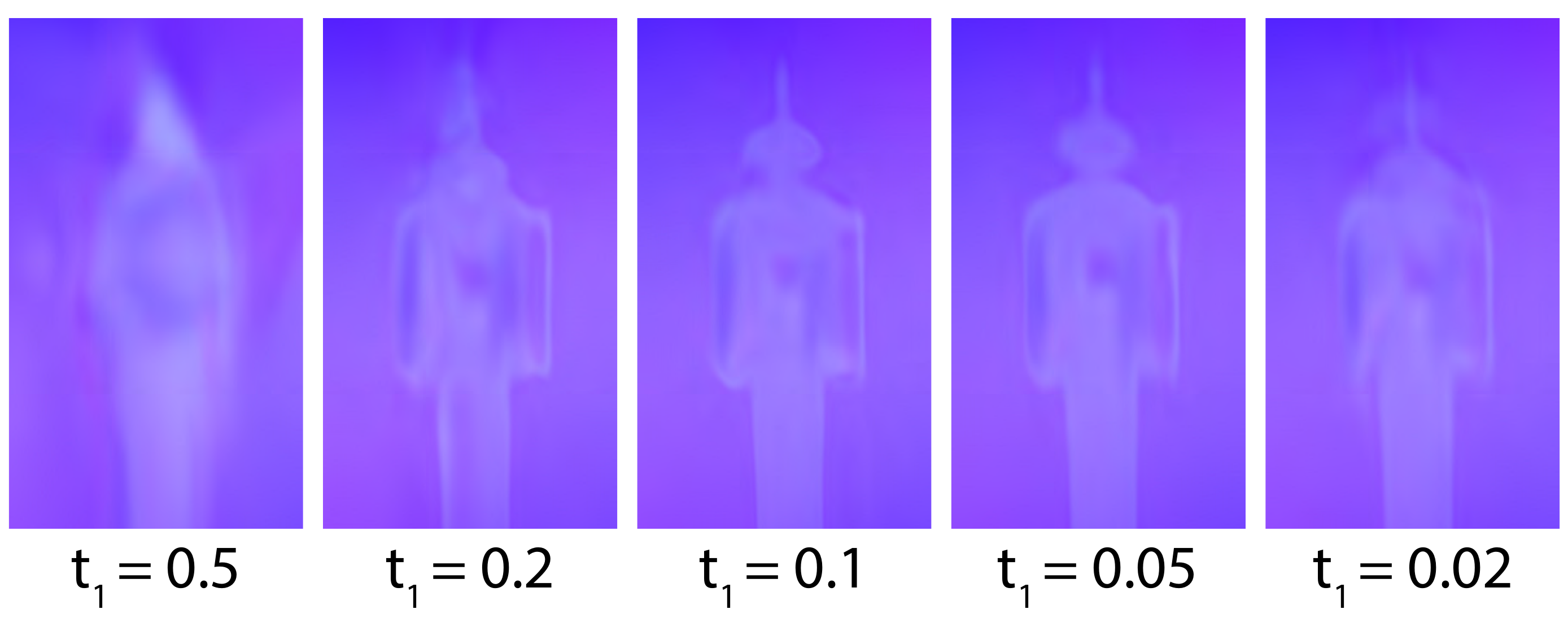}
    \vspace{-0.22in}
    \caption{Effect of changing the distance between the input time coordinates $t_0$ and $t_1$ on the quality of the interpolated flows. In all the cases, $t_0$ is equal to 0.}
    \label{fig:coordablation}
    \vspace{-0.1in}
\end{figure}

\vspace{-0.15in}
\paragraph{Effect of Time Coordinates:} In Fig.~\ref{fig:coordablation}, we show the effect of changing the time coordinates ($t_0$ and $t_1$), which are used as inputs to our hypernetwork. As seen, with a large distance between the two coordinates (0.2 and 0.5), our system is not able to produce high-quality intermediate flows. This is because, in this case, the hypernetwork will not provide sufficient constraints and the estimated coordinate-based network weights  for the two coordinates ($\theta_0$ and $\theta_1$)  become independent. On the other hand, when the coordinates are too close to each other (distance of 0.02), the hypernetwork becomes overly restrictive and cannot estimate proper weights. Distances of 0.1 and 0.05 are ideal and produce the best quality.

\begin{figure}
    \includegraphics[width=\linewidth]{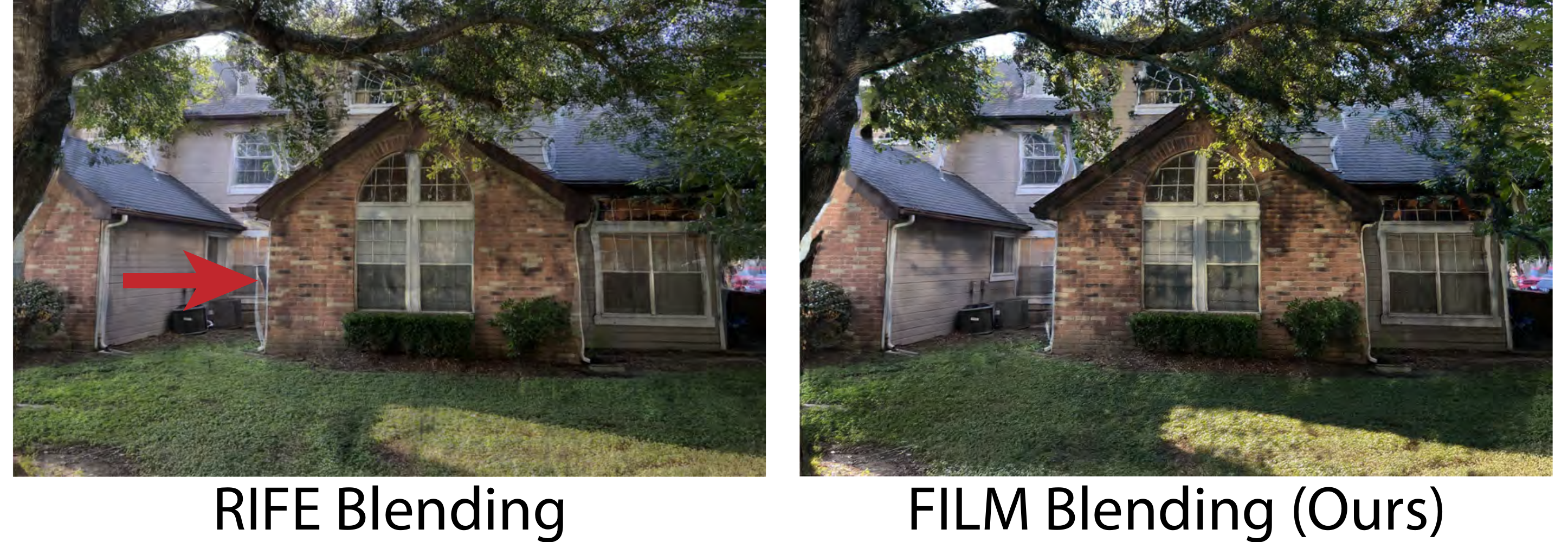}
    \vspace{-0.22in}
    \caption{Comparison between the blending approaches by Huang \etal~\cite{Huang_2022_rife} (RIFE) and Reda \etal~\cite{Reda_2019_CVPR} (FILM) when used with our estimated flows.}
    \label{fig:blendingablation}
    \vspace{-0.2in}
\end{figure}

\vspace{-0.15in}
\paragraph{Effect of Blending:} In our approach, we use Reda~ \etal's blending network~\cite{Reda_2019_CVPR} (FILM). However, we could potentially use the blending network of any other approach that breaks down the process into two stages of flow estimation and blending. In Fig.~\ref{fig:blendingablation}, we compare the quality of the interpolated images using FILM's blending, with that of Huang \etal~\cite{Huang_2022_rife} (RIFE). Note that, in both cases, we use our interpolated flows as the input to their blending system. As seen, while both approaches produce reasonable results, FILM's blending is generally of higher quality thanks to the perceptual Gram matrix loss, used during their training. 

\subsection{Limitations and Future Work}

Our method uses the flows estimated by RAFT~\cite{teed2020raft}, and thus the quality of our results depends on the accuracy of these predicted flows. While RAFT produces high quality flows in a large number of cases, it could potentially fail on challenging scenarios. In these cases, the flow artifacts may appear in our final results. However, since our approach allows us to use any optical flow method, as better flow estimation approaches are developed in the future, we can simply use them to further improve our results.

We also explored the idea of directly interpolating the images, instead of the flows, but were not successful based on our initial experiments. We believe this is because images are significantly more detailed than the optical flows. We leave a thorough investigation of this idea to the future.

\section{Conclusion}

We presented an approach to interpolate between a pair of images of a dynamic scene with lighting variations. We propose to do so by utilizing existing optical flow methods. To calculate the flows between an intermediate frame and the two input images, we interpolate the bidirectional flows estimated using a pre-trained flow network in an implicit manner. Specifically, we encode the bidirectional flows into a coordinate-based network and estimate the flows at any time, by passing the appropriate coordinate as the input. We use the estimated flows within an existing blending network to produce the final interpolated image. We show that our method is able to produce significantly better results than the state of the art on a wide range of challenging scenes.

{\small
\bibliographystyle{ieee_fullname}
\bibliography{egbib}
}

\end{document}